\documentclass{article}
\usepackage{spconf,amsmath,graphicx}
\usepackage{amsfonts}

\title{Adversarial-Prediction Guided Multi-task Adaptation for Semantic Segmentation of Electron Microscopy Images}
\name{Jiajin Yi, Zhimin Yuan, Jialin Peng$^*$}
\address{College of Computer Science and Technology, Huaqiao University, Xiamen, China\\
2004pjl@163.com\thanks{*Corresponding author. This work was supported by NSFC (11771160).}}
\begin{document}
%
\maketitle
\begin{abstract}
Semantic segmentation is an essential step for electron microscopy (EM) image analysis. Although supervised models have achieved significant progress, the need for labor intensive pixel-wise annotation is a major limitation. To complicate matters further, supervised learning models may not generalize well on a novel dataset due to domain shift. In this study, we introduce an adversarial-prediction guided multi-task network to learn the adaptation of a well-trained model for use on a novel unlabeled target domain. Since no label is available on target domain, we learn an encoding representation not only for the supervised segmentation on source domain but also for unsupervised reconstruction of the target data. To improve the discriminative ability with geometrical cues, we further guide the representation learning by multi-level adversarial learning in semantic prediction space. Comparisons and ablation study on public benchmark demonstrated state-of-the-art performance and effectiveness of our approach.
\end{abstract}
\begin{keywords}
Segmentation, domain adaptation, electron microscopy, multi-task learning, adversarial learning
\end{keywords}

\begin{figure*}
\begin{minipage}[b]{1\linewidth}
  \centerline{\includegraphics[width=0.72\textwidth]{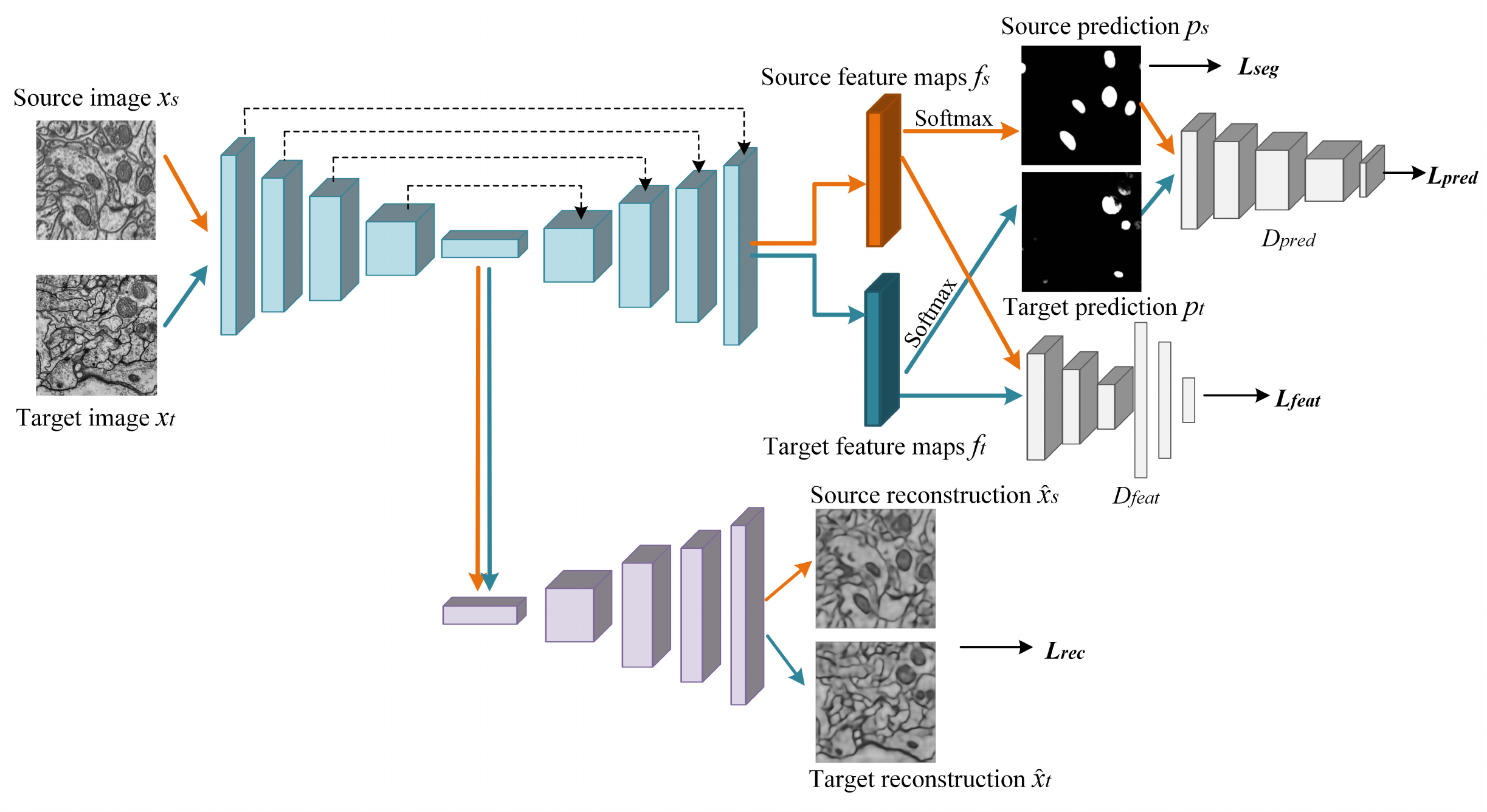}}
\end{minipage}
\caption{The architecture of the proposed APMA-Net for unsupervised domain adaptation. We use the fully convolutional encoder-decoder U-Net \cite{U-Net} as the backbone of segmentation task, and use auto-encoder as the image reconstruction network.  Multiple domain discriminator are applied on both the outputting prediction and decoding feature spaces.  }
\label{fig:workflow}
\end{figure*}

\section{INTRODUCTION}
\label{sec:intro}

The accurate segmentation of electron microscopy (EM) images is an essential step for understanding brain's neuronal structures \cite{bermudez2018domain,peng2019mitochondria}. Current state-of-the-art  methods for medical image segmentation are supervised trained fully convolutional neural networks  in \textit{encoder-decoder} architecture, across which a common obstacle is the severe  dependence on large amounts of pixel-wise labeled data. However, annotating EM images is labor-intensive and time-consuming, which makes it difficult to obtain a large number of labeled EM images. Instead of re-annotating in each domain, generalizing the model trained on the dataset with enough supervising labels (source domain) well to a novel dataset without labels (target domain) is an appealing alternative approach to mitigate the training difficulties of unlabeled dataset.

To reduce domain discrepancy, there have been a lot of studies focusing on unsupervised domain adaptation (UDA) in recent years with the aim to learn domain-invariant representation/model. One frequently-used strategy is to learn domain-invariant features through distribution alignment. For example,  maximum mean discrepancy (MMD) metric and correlation loss were used in \cite{DAN} and  \cite{Coral}, respectively, to match the feature distributions of different domains. In \cite{DANN,ADDA}, a adversarial domain discriminator was introduced to align the features outputted by the feature encoder, which is more effective than directly using statistical metrics \cite{DAN}. In \cite{Rec_Cla_DA}, Ghifary \textit{et al.} proposed to learn a shared feature encoder by augmenting the supervised segmentation task on the source domain with an  auxiliary task for the reconstruction of the unlabeled target data. Similar idea was applied by Roels \textit{et al.} \cite{Y-Net} for EM image segmentation. However, since the decoding representation learning is only guided by labels on source domain, the decoding representation may not generalize well on target domain. Recently, Tsai \textit{et al.} \cite{Outputspace} proposed label-space adaptation based on adversarial learning for multi-label video segmentation. The work is based on the observation that the different domains share similar label spaces, which have rich spatial layouts and local context information in the scenario of multi-class segmentation. However,  for the complicated binary segmentation task, with the label-level adaptation strategy the learned domain-invariant feature representation is less  constrained by the rich visual cues of the images at target domain.  Recently, a semi-supervised domain adaptation method was proposed  in \cite{bermudez2018domain} by applying MMD  at the final stage of the \textit{decoder}. However, this method \textit{cannot} be applied in  our \textit{unsupervised scenario}, since the MMD loss is unsupervised.
Despite these efforts, the performances of UDA are still limited for pixel-wise segmentation tasks with complex image appearance and ambiguous boundaries.

To address these above issues, we propose an Adversarial-Prediction guided Multi-task Adaptation Net (APMA-Net) in the \textit{encoder-decoder} architecture. Given a supervised learned \textit{encoder-decoder} on the source domain, our aim is to adapt it to the target domain  with images having a different visual and structure style. Thereby, we learn shared encoding representations  by seeking better reconstruction of the images of both domains. To improve the discriminative ability of both the encoding and decoding representations on the unlabeled target domain, we further guide the representation learning  process by the similarity in geometric cues  between the two domains, which is captured by multi-level adversarial discriminators in semantic prediction and feature spaces. In this way, the adaptation is jointly guided by both geometrical and visual information, and the distribution alignment happens at multiple representation layers. Compared with  methods \cite{Rec_Cla_DA,Y-Net} only adapting encoding features with auto-encoder, the proposed method adapts to learn domain invariant features more specific to discriminative task on the target domain.

\section{METHOD}
\label{sec:majhead}

Our objective is to learn a sound label predictor for the unlabeled target domain with the help of a labeled source domain. The overview of our proposed adversarial-prediction guided multi-task adaptation net (APMA-Net) is shown in Fig. \ref{fig:workflow}.

To mitigate the domain gap, we perform joint adaptation on two stages, namely encoding stage and decoding stage, in the paradigm of multi-task learning. Concretely, we learn features guided by two tasks: 1) supervised segmentation on the source domain using U-Net \cite{U-Net} as the backbone \textit{encoder-decoder}; 2) unsupervised reconstruction of images on the two domains using an \textit{auto-encoder} following the idea of \cite{Rec_Cla_DA,Y-Net}. The  two tasks share the same encoder. Although the reconstruction branch is unsupervised and not targeted for learning discriminative features, it aims to capture crucial visual and structure cues in the image space of target domain for the adaptation. To make the learned features discriminative on target domain and thus learn good cross-domain label predictor, multi-level domain-adversarial adaptations on decoding feature space and prediction space are conducted on decoding stage. Note that, without the guidance of image reconstruction, the learned features and the cross-domain predictor may contain less information from target image space, which is the most reliable information available about the target domain.

Let the source domain $D_S=\{X_S,Y_S\}$ be a set of labeled  images, and the target domain $D_T=\{X_T\}$ be a novel set of unlabeled images. We define the labeled sample of the source data as $x_s \in X_S$, and the corresponding label as $y_s \in Y_S$. Similarly, the sample of unlabeled target data is defined as $x_t\in X_T$. Moreover, we denote feature maps preceding the output layer of our APMA-Net for source data and target data as $f_s$ and $f_t$, respectively. The final predictions for source and target images $x_s$ and $x_t$ are denoted by $p_s$ and $p_t$, respectively.

\textbf{Encoding stage adaptation.} To learn feature representation capable of encoding enough visual and structure information from both the source  and target domains, we augment an  auto-encoder (denoted as $AE$) to the main segmentation generator network (denoted as $GE$). Our aim is to make the shared encoder of $AE$ and $GE$  learn not only the information of the source domain but also the visual information on the target domain. Specifically, the shared encoder takes both the source data $x_s$ and target data $x_t$ as inputs. When feeding $x_s$ and $x_t$ to the $GE$ branch,  the decoder of $GE$ finally produces  predictions $p_s$ and $p_t$, respectively, and the $AE$ branch produces the reconstructed images $\hat{x}_s$ and $\hat{x}_t$, respectively. It is expected that the reconstructed images $\hat{x}_s$ and $\hat{x}_t$ are close to the corresponding original images $x_s$ and $x_t$ respectively, while the prediction  $p_s$ of $x_s$ is close to the corresponding label $y_s$. We use  cross-entropy loss $L_{seg}$ for source prediction measurement,  and mean squared loss   $L_{rec}$ for the reconstruction measurement, which are defined as follows,
\begin{equation} \label{eq:2}
\mathcal{L}_{seg} = -\mathbb{E}_{(x_{s},y_{s})}[ y_s\mathrm{log}(p_s)+(1-y_s)\mathrm{log}(1-p_s)],
\end{equation}
\begin{equation}
\mathcal{L}_{rec}  = \mathbb{E}_{x_s} [||x_s-\hat{x}_s||^2_2 ] +\mathbb{E}_{x_t}  [||x_t-\hat{x}_t||^2_2 ],
\end{equation}
where  spatial dimensions are omitted   for
simplicity in Eq. (\ref{eq:2}).

Note that the auxiliary reconstruction strategy  as Eq. (2) has been exploited in many tasks including domain adaptation \cite{Rec_Cla_DA,Y-Net}.  In our model, however, we learn to adapt domains simultaneously with the guidance from image space and the guidance from label space.  In this way, we can bias the cross-domain discriminative predictor better to the target domain.

\textbf{Decoding stage adaptation.} The segmentation generator $GE$ is also shared by the two domains (Fig. \ref{fig:workflow}). Taking either the encoded representation of a source image $x_s$ or a target image $x_t$ as input, we can obtain the final feature maps $f_s$, $f_t$ and label predictions $p_s$, $p_t$, respectively. So far, we have just used image information from the target domain and label information from only the source domain to learn compact features, however, we may still lack the ability of discrimination on the target domain. Moreover, the decoding representations in  $GE$ are still less biased to the target domain.

 To improve  discriminative ability of both the encoding and decoding representations on the target domain, we introduce multi-level adversarial learning on both prediction and feature spaces (Fig. \ref{fig:workflow}): 1) at the space of structured prediction, we  leverage a fully convolutional domain discriminator by following the similar idea of \cite{Outputspace}; 2) at the space of  decoding representation, we apply a domain discriminator on the feature maps of the ending decoding layer.

More specifically, we utilize two domain discriminators $D_{pred}$ and $D_{feat}$ for feature alignment at  multi-level prediction spaces. The discriminator $D_{pred}$ is used for the final prediction adaptation (see Fig. \ref{fig:workflow}) and $D_{feat}$ is used for the adaptation of the final decoding representations preceding the output layer. The two discriminators are trained to distinguish domain labels of the inputs from the different domains. With domain-level supervision,  the discriminators $D_{pred}$ and $D_{feat}$ are learned by minimizing the following loss,
\begin{equation}\label{eq:3}
\min_{D_{pred}, D_{feat}}~~-\lambda_{feat}\mathcal{L}_{feat}-\lambda_{pred}\mathcal{L}_{pred},
\end{equation}
\begin{equation}\label{eq:4}
\mathcal{L}_{pred}  =
\mathbb{E}_{x_s}  [\mathrm{log}D_{pred}(p_s)]  +
\mathbb{E}_{x_t} [\mathrm{log}(1-D_{pred}(p_t))],
\end{equation}
\begin{equation}\label{eq:5}
\mathcal{L}_{feat}  =
\mathbb{E}_{x_s}  [\mathrm{log}D_{feat}(f_s)]  +
\mathbb{E}_{x_t} [\mathrm{log}(1-D_{feat}(f_t))],
\end{equation}
in which the image width and height dimensions are   omitted   for
simplicity in Eq. (\ref{eq:4}).
This process of multi-level adversarial learning enables the encoding and decoding representation to be more discriminative on the unlabeled target domain.

The segmentation-reconstruction network is  learned through minimizing  the following loss function,
\begin{equation}\label{eq:6}
\min_{AE, GE}~~
\mathcal{L}_{seg}+\lambda_{rec}\mathcal{L}_{rec}-\lambda_{feat}\mathcal{L}_{feat}^{'}-\lambda_{pred}\mathcal{L}_{pred}^{'},
\end{equation}
in which  $\lambda_{rec}$, $\lambda_{feat}$ and $\lambda_{pred}$ are trade-off weights and
\begin{equation}\label{eq:7}
\mathcal{L}_{feat}^{'}  =
\mathbb{E}_{x_t} [\mathrm{log}(D_{feat}(f_t))], \mathcal{L}_{pred}^{'}  =
\mathbb{E}_{x_t} [\mathrm{log}(D_{pred}(p_t))].
\end{equation}
where we use inverted domain labels to obtain a loss function with lower bound.
To train the proposed model, we iteratively optimize the problem in Eq. (\ref{eq:3}) and Eq. (\ref{eq:6}). Practically, we firstly use the
annotated source images to train an initial $GE$.


\section{RESULTS}
\label{sec:majhead}
 Given an annotated domain, our evaluation task is  unsupervised  segmentation of mitochondria in EM images from a novel domain with severe domain shift. We use the well annotated EPFL  Data\footnote{https://cvlab.epfl.ch/data/em} as the source domain, which is  scanned by Focused Ion Beam Scanning EM (FIB-SEM). It is an image stack of size 165 $\times$ 1024 $\times$ 768 taken from CA1 hippocampus region of a mouse brain.  For target domain dataset, we use an image stack of size 20 $\times$ 1024 $\times$ 1024 \cite{targetdata}, which is acquired by serial section Transmission EM (ssTEM). This image stack is taken from  the Drosophila melanogaster third instar larva Ventral Nerve Cord. Since the target domain dataset contains a small number of serial sections, we split the target dataset along the $x$ axis, with 67$\%$  for training and 33$\%$ for testing.
\begin{figure}[t]
\includegraphics[width=0.48\textwidth]{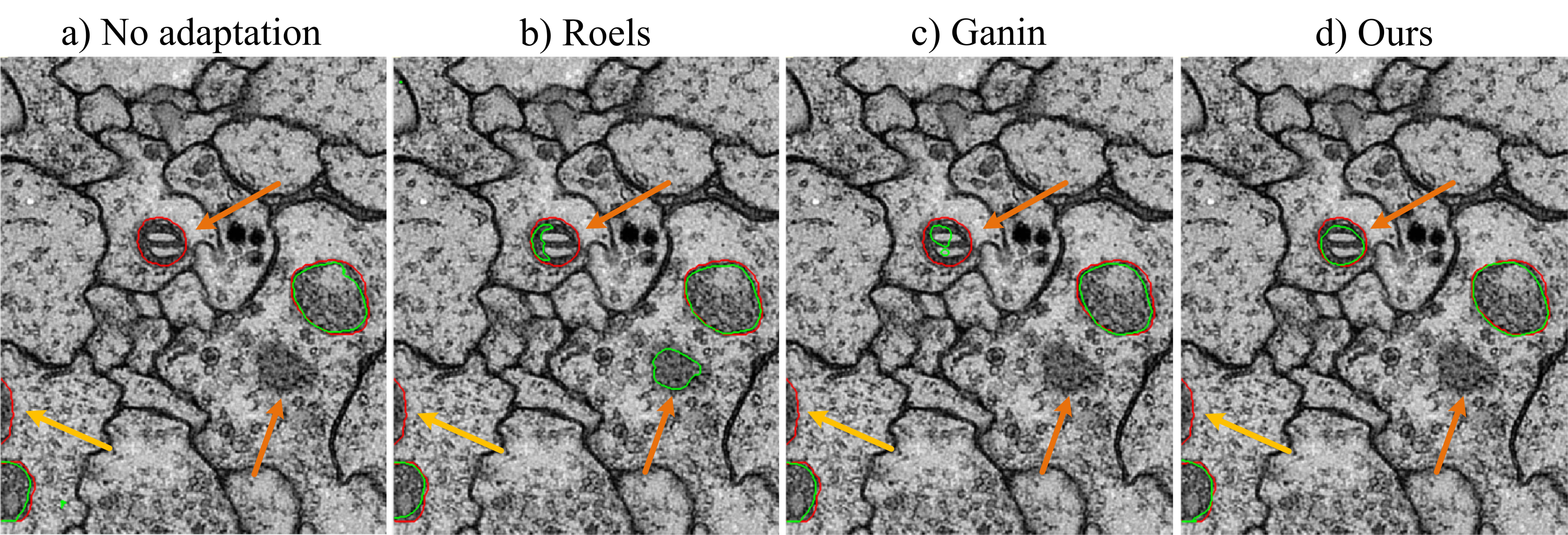}%
\caption{Visual comparison of different segmentation results.}
\label{fig:results}
\end{figure}

We implement our network using PyTorch  on a 1080Ti GPU.  For our proposed network, we use the Adam \cite{kingma2014adam} optimizer with the learning rate as 2$\times$$10^{-4}$ and polynomial decay with a power of 0.9. The trade-off parameters are set as $\lambda_{rec}$=$10^{-3}$, $\lambda_{feat}=\lambda_{pred}=1\times 10^{-3}$.

 We compare our method with
1) No adaptation that uses a U-Net trained only on the source domain to segment the target domain; 2) Roels \cite{Y-Net} which aligns the encoder features of source and target domains by adding a reconstruction decoder to the U-Net; 3) Ganin \cite{DANN} that learns domain-invariant feature  by using an adversarial domain discriminator.

Figure \ref{fig:results} shows visual comparison results, in which  falsely detected regions and missing regions by No adaptation, Roels \cite{Y-Net} and/or Ganin \cite{DANN}  are highlighted with orange arrows.  Our method shows obviously reduced detection error. However, there are also regions (highlighted with yellow arrow)  that are failed to be detected by all methods.  The quantitative  results are shown in Table \ref{tab:1}, which are evaluated using Jaccard index (JAC) and Dice similarity coefficients (DSC). Our proposed method yields an accuracy of 53.6$\%$ in JAC and 69.8$\%$ in DSC, respectively, which is significantly better than other unsupervised domain adaptation methods.

\begin{table}[t]
\caption{Comparison with state-of-the-art methods for unsupervised mitochondria segmentation in EM images.}
\centering
\label{tab:1}      
 \setlength{\tabcolsep}{3mm}
\begin{tabular}{lcc}
\hline\noalign{\smallskip}
Experiments & DSC($\%$) & JAC($\%$)  \\
\noalign{\smallskip}\hline\noalign{\smallskip}
No adaptation	&45.3	& 29.3\\
Roels \cite{Y-Net} &	66.4	& 49.7\\
Ganin \cite{DANN}		 & 66.7 &  50.0\\
Our APMA-Net	& 69.8	& 53.6 \\
\noalign{\smallskip}\hline
\end{tabular}
\end{table}
To validate the effectiveness of our joint encoding and decoding adaptation approach, five ablated versions of our model have been compared: 1) only ENcoding stage adaptation with Auto-Encoder ($EN$); 2) only the final DEcoding features adap{}tation (${DE}_{feat}$); 3) only the  DEcoding prediction adaptation (${DE}_{pred}$); 4) the combination of 1) and 2) ($EN$+$DE_{feat}$);  5) our final model (APMA-Net).  The evaluation results are shown in Table \ref{tab:2}. In comparison with No adaptation, it is obviously that the method with the $EN$  outperforms  it by a large margin, which indicates that the learned label predictor is more biased towards the target domain.  The performances of ${DE}_{feat}$ and ${DE}_{pred}$ are both superior than No adaptation and $EN$, which indicates that the effectiveness of the prediction space adaptation. Furthermore, the combination of ${DE}$ and ${DE}_{feat}$ can obviously perform better than using them separately, which is further improved by adding ${DE}_{pred}$  (i.e. APMA-Net). All above results show that the adaptation which is guided jointly by label domain and image information is an effective way to mitigate the domain gap.

\section{CONCLUSION}
\label{sec:majhead}
In this paper, we proposed a multi-task adaptation method to address the problem of unsupervised segmentation of EM images. To improve the discriminative ability of the cross-domain label predictor on the unlabeled domain,
we adopted the multi-level adversarial learning in semantic prediction space to leverage domain-level label information.
Experimental results showed that our proposed method can achieve state-of-the-art performance in accuracy and visual quality.

\begin{table}[!t]
\caption{Ablation study of our APMA-Net. $EN$:  encoding stage adaptation by auto-encoder; $DE_{feat}$: decoding stage adaptation by aligning features; $DE_{pred}$: decoding stage adaptation by aligning predictions.}
\centering
\label{tab:2}
\begin{tabular}{lcc}
\hline\noalign{\smallskip}
Experiments & DSC($\%$) & JAC($\%$)  \\
\noalign{\smallskip}\hline\noalign{\smallskip}
$EN$ & 	66.4	&49.7\\
$DE_{feat}$ & 	67.6&	51.0\\
$DE_{pred}$ & 	67.4	& 50.8\\
$EN$+$DE_{feat}$	&68.3&	51.8\\
$EN$+$DE_{feat+pred}$ (APMA-Net) &	69.8&	53.6\\
\noalign{\smallskip}\hline
\end{tabular}
\end{table}


\end{document}